\documentclass{article}
\usepackage{ijcai20}

\usepackage{times}

\usepackage{soul}
\usepackage{url}
\usepackage[utf8]{inputenc}
\usepackage[small]{caption}
\usepackage{graphicx}
\usepackage{amsmath}
\usepackage{booktabs}
\usepackage[symbol]{footmisc}

\usepackage{xcolor}

\title{Privacy-Preserving Technology to Help Millions of People: \\
Federated Prediction Model for Stroke Prevention}

\author{
Ce Ju$^{1,}$\footnote{Equal Contribution}\and
Ruihui Zhao$^{2,*}$\and
Jichao Sun$^{2,*}$\and
Xiguang Wei$^{1,*}$\and \\
Bo Zhao$^2$\and
Yang Liu$^1$\and
Hongshan Li$^3$\and
Tianjian Chen$^1$\and \\
Xinwei Zhang$^4$\and 
Dashan Gao$^{5, 6}$\and
Ben Tan$^1$\and
Han Yu$^7$\and
Chuning He$^8$\And
Yuan Jin$^8$
\affiliations
$^1$WeBank Co., Ltd., AI Department \hspace*{0.5em} $^2$Tencent Jarvis Lab\\
$^3$Tsinghua-Berkeley Shenzhen Institute, Tsinghua University \hspace*{0.5em} $^4$Lehigh University\\
$^5$Hong Kong University of Science and Technology \hspace*{0.5em} $^6$Southern University of Science and Technology\\
$^7$Nanyang Technological University \hspace*{0.5em} $^8$ Shenzhen Gradient Technology Co.,Ltd.
\emails
\{ceju, xiguangwei, yangliu, tobychen, btan\}@webank.com\\
\{zacharyzhao, jichaosun, kevinbzhao\}@tencent.com\\
lhs17@mails.tsinghua.edu.cn, xiz415@lehigh.edu, dgaoaa@connect.ust.hk,\\
han.yu@ntu.edu.sg, 
chuning.he@gradient.healthcare,
yuan.jin@gradient.healthcare\\
}

\begin{document}

\maketitle

\begin{abstract}

Prevention of stroke with its associated risk factors has been one of the public health priorities worldwide. Emerging artificial intelligence technology is being increasingly adopted to predict stroke. Because of privacy concerns, patient data are stored in distributed electronic health record (EHR) databases, voluminous clinical datasets, which prevents patient data from being aggregated and restrains AI technology to boost the accuracy of stroke prediction with centralized training data. In this work, our scientists and engineers propose a privacy-preserving scheme to predict the risk of stroke and deploy our federated prediction model on cloud servers. Our system of federated prediction model asynchronously supports any number of client connections and arbitrary local gradient iterations in each communication round. It adopts federated averaging during the model training process, without patient data being taken out of the hospitals during the whole process of model training and forecasting. With the privacy-preserving mechanism, our federated prediction model trains over all the healthcare data from hospitals in a certain city without actual data sharing among them. There- fore, it is not only secure but also more accurate than any single prediction model that trains over the data only from one single hospital. Especially for small hospitals with few confirmed stroke cases, our federated model boosts model performance by $10\% \sim 20\%$ in several machine learning metrics. To help stroke experts comprehend the advantage of our prediction system more intuitively, we developed a mobile app that collects the key information of patients’ statistics and demonstrates performance comparisons between the federated prediction model and the single prediction model during the federated training process.
\end{abstract}

\section{Introduction}
\begin{figure}[h]
	\centering
	\includegraphics[width=0.40\textwidth]{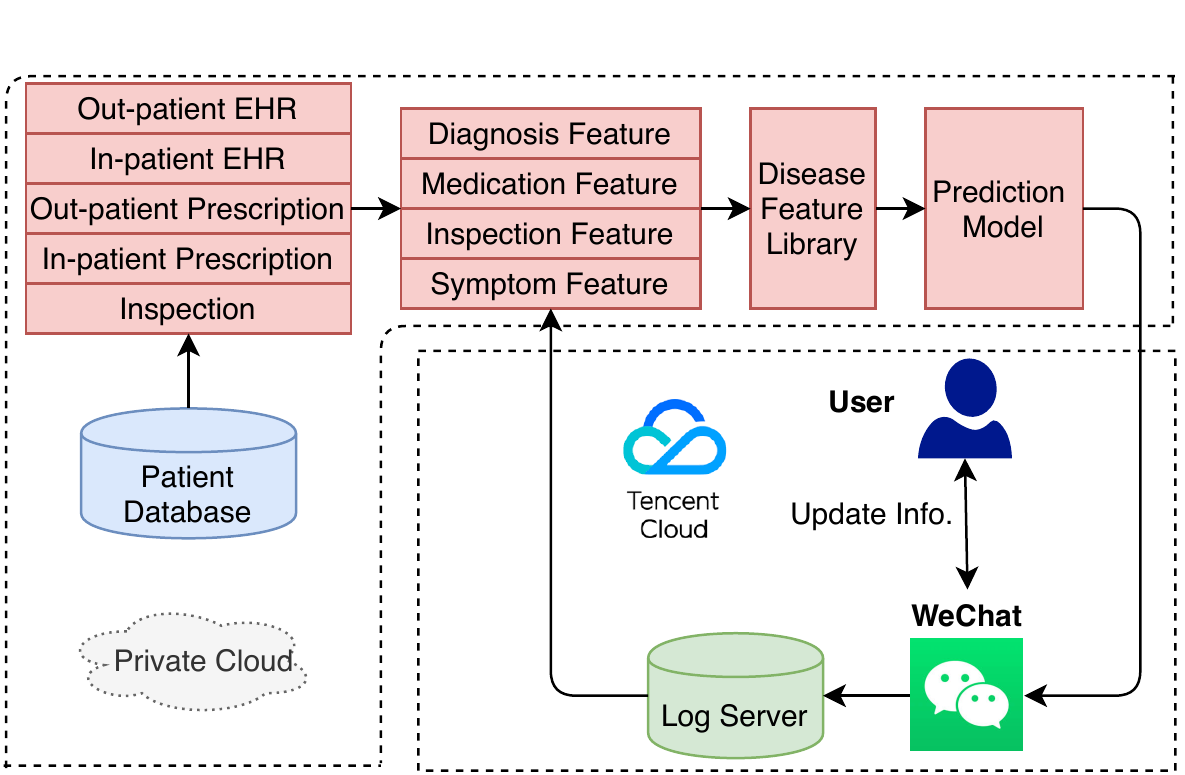}
	\caption{Workflow of the Stroke Prediction Model}
	\vspace{-1mm}
	\label{f1}
\end{figure}

Recent data privacy concerns begin with an investigation of Facebook’s data privacy violations in collecting user information for uninformed use~\cite{tuttle2018facebook}. Hence, acts like the General Data Protection Regulation (GDPR) are proposed to prohibit organizations from exchanging data without explicit user approval~\cite{regulation2016regulation}. Data, collected by health care providers, include a record of clinical metrics along with information pertinent to health and wellness. Although the use of big data in healthcare advances and gains popularity, it may lead to severe data privacy violations~\cite{jiang2017artificial,miotto2018deep}. Therefore, it is crucial to design and implement a privacy-preserving mechanism for AI technology in the field of healthcare. We adopt an emerging and powerful technique, federated learning framework~\cite{DBLP:journals/corr/McMahanMRA16,DBLP:journals/corr/KonecnyMRR16}, to address the data privacy because it can jointly train a machine learning model using data from different clients without the need for actually sharing data between clients. The federated model performs not only more accurate but also more secure than the traditional statistical model in medical research~\cite{bender2005generating}.

In this work, Tencent and WeBank collaboratively develop a privacy-preserving technology for the prediction of stroke. Tencent’s tech-experts have successfully developed a prediction model for stroke with advanced AI technology supported by Tencent’s cloud server in the past year. To solve the scalability problem due to privacy concerns, We- Bank’s tech-experts help establish and implement a federated version for Tencent’s prediction model based on their self- developed privacy-preserving framework with FATE~\footnote{FedAI (https://www.fedai.org/) is a community that helps businesses and organizations build AI models effectively and collaboratively, by using data in accordance with user privacy protection, data security, data confidentiality, and government regulations.}, an industrial level open-source secure computing framework. Our privacy-preserving framework makes the impossible practical problem possible. Combing advanced AI technology and great sensitivity to privacy protection, our privacy-preserving framework works both accurately and securely. This stroke prediction application in healthcare is the second project hit- ting the ground for the new privacy-preserving technology yet proposed by WeBank from China, which may help millions of people to prevent stroke.
\section{Related Work}
Federated learning is a machine learning setting in which clients solve a machine learning problem through collaboration~\cite{kairouz2019advances,yang2019federated}. During the model training, the data never leave the database of clients. Because health care data are strictly regulated by data compliance rules in China, clinical data in hospitals are very appropriate for this setting. Existing literature demonstrates the great potential of this approach in health care. Brisimi\cite{brisimi2018federated} developed an iterative cluster Primal-Dual Splitting (cPDS) algorithm for patients EHR in a decentralized fashion. Sheller~\cite{sheller2018multi} introduced the first use of federated learning in the medical image for multi-institutional collaboration. Li~\cite{li2019privacy} investigated the feasibility of applying differential-privacy techniques to protect the patient data in a federated learning setup for brain tumor segmentation. Ju and Gao~\cite{gao2019hhhfl,ju2020federated} are first to investigate the feasibility of the federated learning framework to enable a distributed training of brain-computer interface models from multiple datasets with heterogeneous configurations.

\section{Methodology}
\subsection{Workflow of the Stroke Prediction Model}
Figure \ref{f1} illustrates workflow of the stroke prediction model.
\subsubsection{Data Preprocessing}
Our tech-experts clean and normalize raw data of patients collected from hospitals in a city of Central China with medical knowledge and AI techniques, such as natural language processing and relation networks, into structured clinical data. Key information of these structured clinical data includes out-patient/in-patient EHRs, out-patient/in-patient prescriptions, and inspections, which are distributed stored in each hospital’s private cloud. Each hospital’s private cloud is not permitted to communicate with each other. Furthermore, the structured clinical data is extracted to model features and grouped in a disease feature library in a private cloud.
\subsubsection{Missing Data Imputation}
An omission in clinical data often yields an inaccuracy in a prediction model. Data imputation is typically used to remedy the omission~\cite{efron1994missing}. We fill-in missing entries using the following methods:
\begin{itemize}
\item Column mean: replace the missing value with the mean of non-missing values.
\item Column median: replace the missing value with the me- dian of non-missing values.
\end{itemize}
We rounded the imputed value as the nearest discrete value after the above post-processing step.
\subsubsection{Feature Selection}
Because only a small number of attributes in clinical data is highly relevant to stroke, our medical experts manually select 119 model features based on risk factors analyzed by medical studies, including systolic blood pressure, use of antihypertensive therapy, prior coronary disease, etc.~\cite{chambless2004prediction,lumley2002stroke,voko2004american}
\subsubsection{Prediction Classifier}
Our prediction classifier for the risk of stroke, outputting the risk score, is established on the model features in the feature library. We pick the following approaches as the prediction classifier:
\begin{itemize}
\item Logistic Regression~\cite{kleinbaum2002logistic}
\item Random Forest~\cite{liaw2002classification}
\item XGBoost~\cite{chen2016xgboost} 
\item Neural Networks~\cite{lecun2015deep}
\end{itemize}
The computing platform for AI is supported by our Tencent Cloud server~\footnote{Tencent Cloud (https://intl.cloud.tencent.com/) is a secure, reliable and high-performance cloud compute service provided by Tencent.}. The risk score can be found in a mini-program on the instant messaging WeChat~\footnote{WeChat (https://www.wechat.com/en/) is a Chinese multi-purpose messaging, social media and mobile payment app developed by Tencent.}. The results of the medical test for patients will form a feedback loop in our workflow to upgrade our feature library and prediction model. 

\begin{figure}
	\centering
	\includegraphics[width=0.49\textwidth]{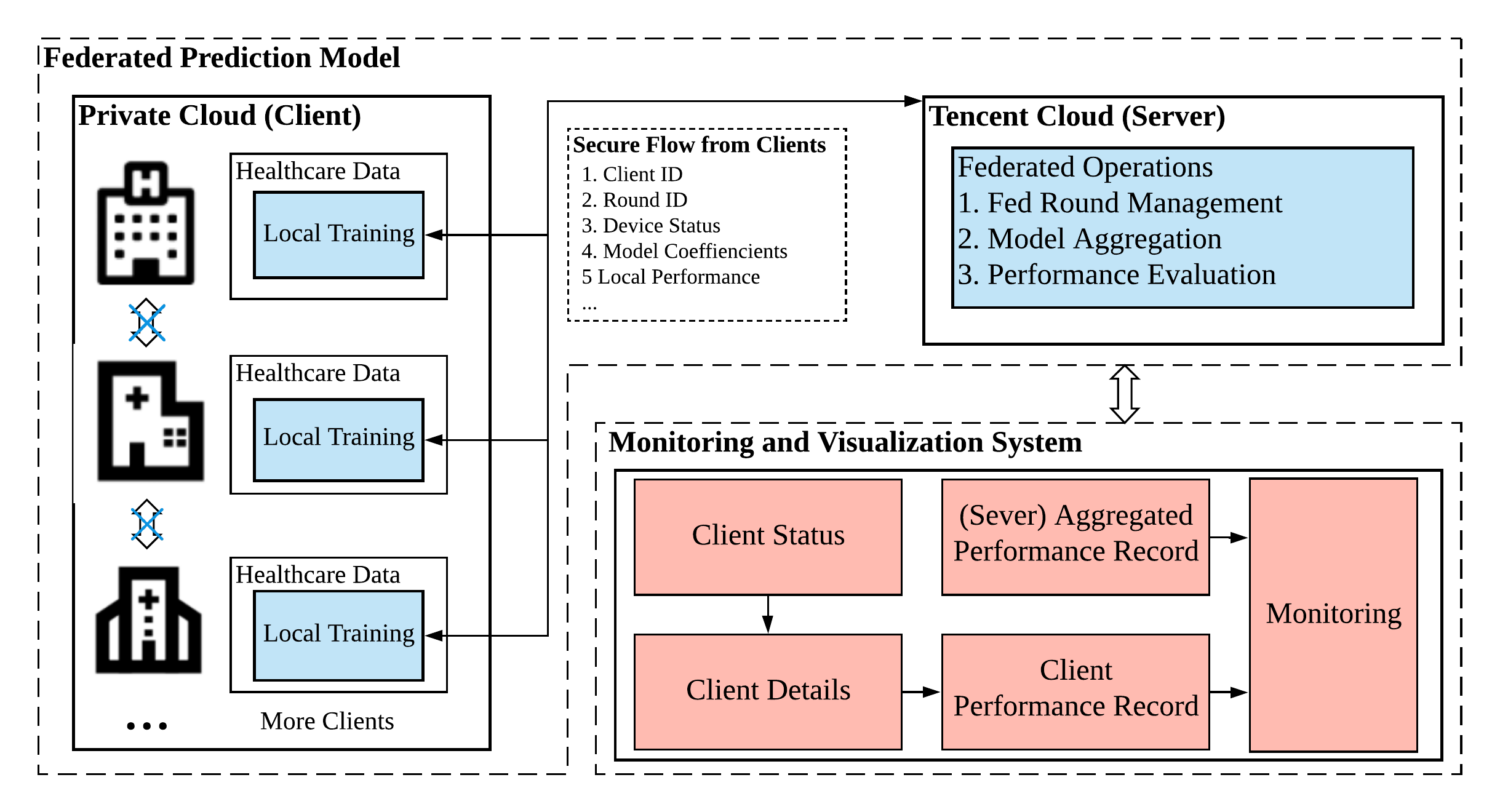}
	\caption{Illustration of Architecture of Federated Prediction Model}
	\vspace{-1mm}
	\label{f2}
\end{figure}

\begin{table}
\begin{center}
\begin{tabular}{| l | l | l | l |}
\hline
Experimental Setting & AUC Mean & AUC Std.  \\ \hline
Hospital A Local Training& 0.812 &  0.027  \\ \hline 
Hospital B Local Training& 0.799 &  0.062  \\ \hline
{Hospital C Local Training}& 0.797 &  0.067  \\ \hline
{Hospital D Local Training}& 0.720 &  0.035  \\ \hline
{Hospital E Local Training}& 0.701 &  0.026  \\ \hline \hline
{Federated Training}& 0.813 &  0.018  \\ \hline 
{Centralized Training}& 0.814 &  0.014  \\ \hline
\end{tabular}
\end{center}
\caption{Cross-validation accuracy of stroke prediction model on different experimental settings.\label{table1}}
\end{table}

\subsection{Federated Prediction Model}\label{FPM}
Our federated prediction system adopts the baseline method, federated averaging~\cite{DBLP:journals/corr/McMahanMRA16}, during model training. Federated averaging is conducted on each client’s (i.e. hospital) private cloud for feature mapping aggregation and classifier aggregation. Specifically, as illustrated in the blue part of Figure~\ref{f2}, the techniques of federated learning follow a server-client setting. The Tencent Cloud sever acts as a model aggregator. In each round $t$, the server $i$ collects updated feature mapping models with model weights $w_t^i$ and classifiers from each client for model aggregation as follows,
\[
w_{t+1}\leftarrow \frac{1}{m} \cdot \sum_{i=1}^m w_t^i.
\]
After model aggregation, the Tencent Cloud server sends the updated global model to each client. When a client receives the model sent by the server, it updates the model with its lo- cal data in the private cloud. The training process continues until the model converges. Our system of federated prediction model is asynchronous supporting any number of client connections and any local gradient iteration in each communication round.

\subsection{Monitoring and Visualization System}

\begin{figure}
\begin{tabular}{c c}
\begin{minipage}{3.8cm}
\centering
\includegraphics[width=3.8cm]{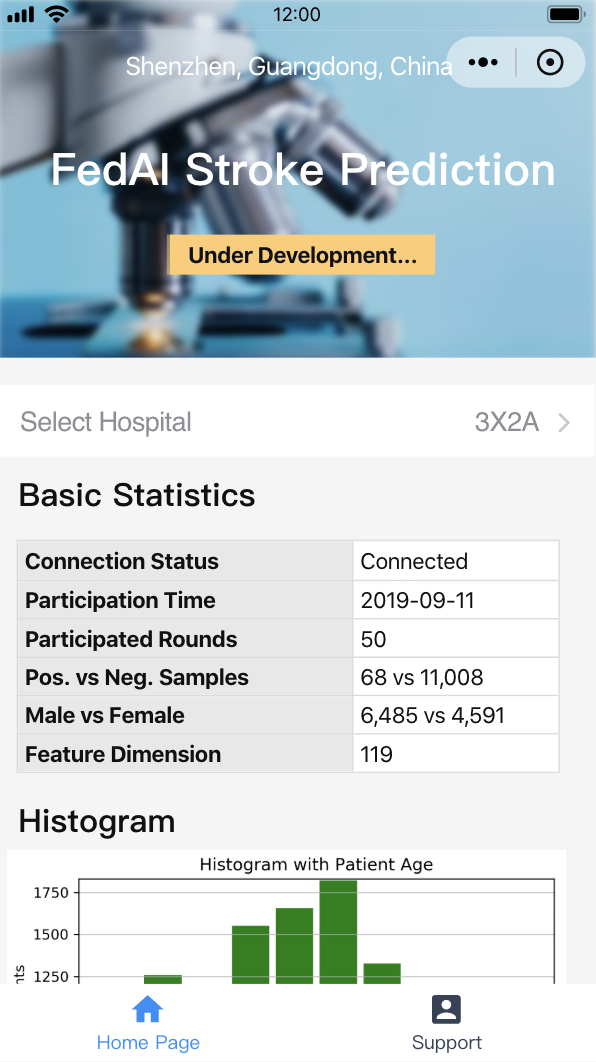}
\end{minipage}
&
\begin{minipage}{3.8cm}
\centering
\includegraphics[width=3.8cm]{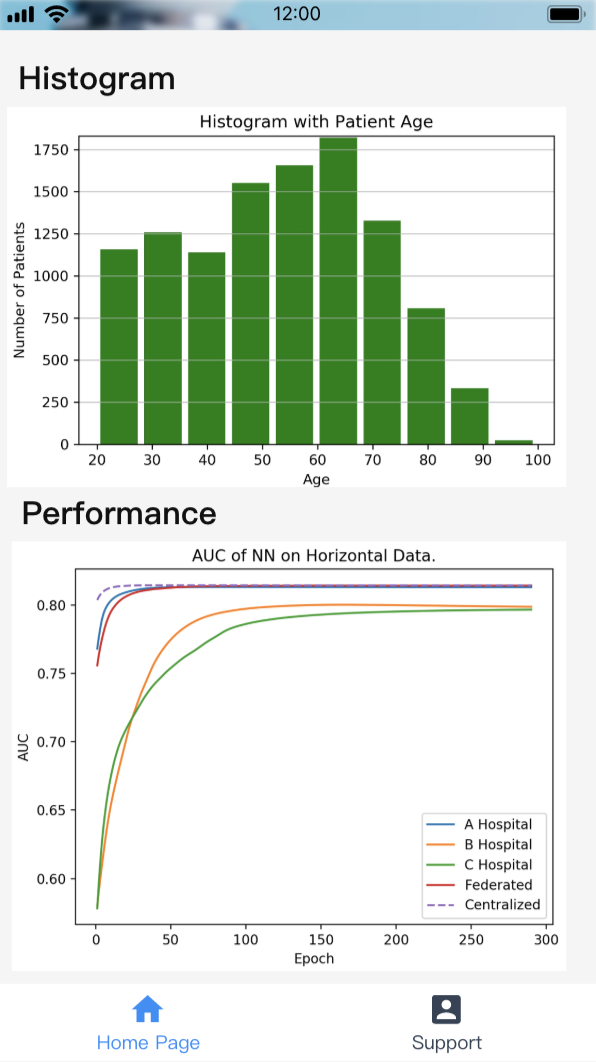}
\end{minipage}
\end{tabular}
\caption{Illustration of a Mini-program FedAI Stroke Prediction.}
\vspace{-1mm}
\label{f3}
\end{figure}

We develop a monitoring and visualization system as shown in the red right part of Figure~\ref{f2} to better explain and control the federated training process by detecting the data flow through clients and the sever. The user interface for the clients in our system, a mini-program called FedAI Stroke Prediction, is illustrated in Figure \ref{f3}. On the first page of the user interface, each user, doctors in hospitals, can view the basic statistics of patients, including numbers of male and female, numbers of positive and negative samples, communication rounds, feature dimensions in model and others. There are also histograms for several patient statistics (i.e. patient age). The performance, such as the area under the curve (AUC) score, between the federated prediction model and the single prediction model, is plotted on the second page of our user interface.

\section{Experimental Analysis}
In this part, we evaluate our federated prediction model in the hospitals in a city of Central China. Hospitals A, B, and C are Top 3 3A hospitals with $3\% \sim4\%$ clinically confirmed stroke patients. The confirmed ratio in Hospital D or E is less than $1\%$. The stroke prediction model is a 3-layer neural network classifier. We conduct 5 experiments to calculate the standard error of the AUC score. The test set for each experiment is fixed and sampled from patients in the whole city. From Table \ref{table1}, we find that the model’s AUC mean under the federated training experiment is almost the same as the one under the centralized training experiment with a little bigger variance. Because the number of patients in Hospital A makes up around $50\%$ of the total patient number, the model performance of Hospital A dominates in the joint training, close to the one of centralized training experiments. With the privacy-preserving mechanism, our federated prediction model trains over all the healthcare information from hospitals in a city and works better than any single prediction model that trains over the information only in one single hospital, especially the small hospital with few confirmed cases, increased by $10\%\sim 20\%$ in model evaluation metric AUC score.

\section{Discussion}
Overall, in this work, we design a privacy-preserving mechanism to boost the performance of the prediction model for stroke and establish the first hitting-ground privacy-preserving pipeline in healthcare worldwide, which has pro- found social value.

\bibliographystyle{named}
\bibliography{refs}

\end{document}